\documentclass{article} 
\usepackage[preprint]{colm2026_conference}

\usepackage{microtype}
\usepackage{hyperref}
\usepackage{url}
\usepackage{booktabs}

\usepackage{amssymb}
\usepackage{amsmath}
\usepackage{amsthm}
\usepackage{multirow}
\usepackage{graphicx}
\usepackage{wrapfig}
\usepackage{algorithm}
\usepackage{algpseudocode}
\usepackage{pdflscape}
\usepackage{rotating}
\usepackage{subcaption}

\newtheorem{proposition}{Proposition}

\usepackage{lineno}

\definecolor{darkblue}{rgb}{0, 0, 0.5}
\hypersetup{colorlinks=true, citecolor=darkblue, linkcolor=darkblue, urlcolor=darkblue}

\title{Beyond Surface Statistics: Robust Conformal Prediction for LLMs via Internal Representations}

\author{Yanli Wang \\
Imperial College London \\
\texttt{yanli.wang22@imperial.ac.uk}
\AND
Peng Kuang \\
Zhejiang University \\
\texttt{22321067@zju.edu.cn}
\AND
Xiaoyu Han \\
University of Illinois Urbana-Champaign \\
\texttt{xiaoyu27@illinois.edu}
\AND
Kaidi Xu \\
City University of Hong Kong \\
\texttt{kaidixu@cityu.edu.hk}
\AND 
Haohan Wang \\
University of Illinois Urbana-Champaign \\
\texttt{haohanw@illinois.edu}
}

\begin{document}

\ifcolmsubmission
\linenumbers
\fi

\maketitle

\begin{abstract}
Large language models are increasingly deployed in settings where reliability matters, yet output-level uncertainty signals such as token probabilities, entropy, and self-consistency can become brittle under calibration--deployment mismatch. Conformal prediction provides finite-sample validity under exchangeability, but its practical usefulness depends on the quality of the nonconformity score. We propose a conformal framework for LLM question answering that uses internal representations rather than output-facing statistics: specifically, we introduce Layer-Wise Information (LI) scores, which measure how conditioning on the input reshapes predictive entropy across model depth, and use them as nonconformity scores within a standard split conformal pipeline. Across closed-ended and open-domain QA benchmarks, with the clearest gains under cross-domain shift, our method achieves a better validity--efficiency trade-off than strong text-level baselines while maintaining competitive in-domain reliability at the same nominal risk level. These results suggest that internal representations can provide more informative conformal scores when surface-level uncertainty is unstable under distribution shift.
\end{abstract}

\section{Introduction}
Large language models (LLMs) are increasingly deployed in settings where reliability matters, from general question answering and decision support to higher-stakes domains such as law, finance, and medicine, where users care not only about average accuracy but also about whether a model's output should be trusted \citep{pmlr-v235-huang24x, machcha2026knowingabstainmedicalllms}. Yet uncertainty quantification for LLM generation remains difficult. Common confidence proxies based on token probabilities, entropy, or self-consistency can become brittle under distribution shift, precisely when deployment risk is highest \citep{kuhn2023semanticuncertaintylinguisticinvariances}. Moreover, because many surface forms can express the same meaning, output-level uncertainty need not align with uncertainty over the underlying semantic decision.

Conformal prediction (CP) is appealing because it wraps arbitrary predictors with finite-sample validity guarantees under exchangeability \citep{angelopoulos2022gentleintroductionconformalprediction}. In LLM deployment, however, calibration and test data often differ across domains, topics, and prompting styles, so this guarantee can degrade sharply \citep{NEURIPS2021_0d441de7}. Shift-aware conformal methods can help, but they typically require informative covariates that partition the input space or accurate importance weights that characterize the calibration--test shift \citep{tibshirani2020conformalpredictioncovariateshift, barber2023conformalpredictionexchangeability}. For LLMs, deriving such structure from text alone is difficult: prompt similarity and lexical overlap are often only shallow proxies for the latent factors that govern reliability. This suggests that the bottleneck is not only how CP is adapted to shift, but also which uncertainty signal is being conformalized.

Recent conformal methods for LLMs make this limitation especially clear. API-only approaches conformalize black-box uncertainty signals without logit access \citep{su2024apienoughconformalprediction}; sampling-based methods build correctness-oriented uncertainty sets from multiple generations \citep{wang2024conuconformaluncertaintylarge}; and selective-answering methods calibrate thresholds to control downstream risk for a single returned answer \citep{wang2025coinuncertaintyguardingselectivequestion}. Even domain-shift-aware methods for LLMs still rely mainly on surface representations to assess similarity or reweight calibration data \citep{lin2025domainshiftawareconformalpredictionlarge}. Across these settings, the dominant signals remain output-facing \citep{quach2024conformallanguagemodeling}, so when reliability-relevant shift is not captured by observable surface features, these methods may inherit the fragility of text-level statistics.

A complementary line of work suggests that reliability signals may reside inside the model rather than in the final output alone. Prior studies show that LLM internal representations preserve semantic and reliability-relevant structure that is only partially visible from decoded text or final-layer statistics \citep{azaria-mitchell-2023-internal, chen2024insidellmsinternalstates}. Recent layer-wise analyses further suggest that hallucinations and unanswerable cases manifest as information deficiency or instability across depth, and that aggregating evidence over layers can be more informative than probing only the final layer \citep{kim2025detectingllmhallucinationlayerwise}. These observations motivate a conformal perspective built directly on internal representations.

In this work, we operationalize that perspective within a standard conformal pipeline for LLM question answering. We introduce \emph{Layer-wise Information} (LI) scores 
, computed from how input context reshapes predictive entropy across model depth, and use them as nonconformity measures in split conformal prediction. The conformal wrapper itself is unchanged: our contribution is to replace an output-level uncertainty score with an internal, answer-level reliability score aggregated from hidden-state trajectories across sampled candidate answers. Accordingly, we do not claim that internal representations remove the need for conformal assumptions or restore formal validity under domain shift. Our claim is narrower and empirical: if layer-wise information ranks admissible answers more faithfully than output-level scores, then the same conformal wrapper can yield better validity--efficiency trade-offs, especially when calibration and deployment domains differ.

Our contributions are threefold: (1) we propose LI-based nonconformity scores that move conformal uncertainty estimation from output-level statistics to internal layer-wise signals; (2) across closed-ended, open-domain, and cross-domain QA benchmarks, we show that internal-score conformalization achieves a stronger empirical validity--efficiency trade-off than baselines based on API-only, sampling-based, and selective-answering uncertainty measures, with the clearest gains under cross-domain shift; and (3) we position internal representations as a practical interface between mechanistic reliability signals in LLMs and conformal uncertainty quantification, highlighting a path toward more robust conformal scoring beyond the calibration distribution.

\section{Related Work}
\paragraph{Conformal uncertainty quantification for LLMs.}
Conformal prediction (CP) provides finite-sample guarantees under exchangeability and is now a standard tool for uncertainty quantification \citep{angelopoulos2022gentleintroductionconformalprediction, angelopoulos2026theoreticalfoundationsconformalprediction}. A broad literature studies how CP behaves beyond the classical exchangeable setting, including adaptive conformal inference under distribution shift, covariate-shift-aware reweighting, and more general analyses beyond exchangeability \citep{NEURIPS2021_0d441de7, tibshirani2020conformalpredictioncovariateshift, barber2023conformalpredictionexchangeability}. Related work also develops conditional or approximate-conditional guarantees and risk-control formulations beyond standard marginal coverage \citep{10.1093/jrsssb/qkaf008, plassier2024probabilisticconformalpredictionapproximate}. These ideas have recently been adapted to LLMs and language generation, from early work on closed-ended and multi-choice QA \citep{kumar2023conformalpredictionlargelanguage} to open-ended generation, API-only settings, and correctness-oriented uncertainty sets for free-form QA \citep{quach2024conformallanguagemodeling, su2024apienoughconformalprediction, wang2024conuconformaluncertaintylarge}. Other lines study factuality, long-form generation, and selective or abstaining deployment, including COIN and SConU, which calibrate thresholds or detect uncertainty outliers to improve robustness in QA settings \citep{mohri2024languagemodelsconformalfactuality, NEURIPS2024_d02ff1ae, wang2025coinuncertaintyguardingselectivequestion, wang-etal-2025-sconu}. Despite differences in interface and objective, most existing methods remain output-facing, deriving nonconformity or confidence from final-output statistics. Our work is similar in goal but different in mechanism: instead of proposing another output-level proxy, we study whether internal layer-wise scores can better support conformal prediction for LLMs.

\paragraph{Internal representations as reliability signals.}
A complementary line of work suggests that the most informative reliability signals may lie in the model’s internal representations rather than in decoded outputs alone. Early evidence showed that hidden activations can reveal latent knowledge and truthfulness signals that are only weakly reflected in surface probabilities or generated text \citep{azaria-mitchell-2023-internal, burns2024discoveringlatentknowledgelanguage}. This view is reinforced in hallucination detection: \citet{chen2024insidellmsinternalstates} show that internal states retain substantial detection power even when output-level statistics are weak, and related activation-based approaches likewise probe internal computation rather than final responses alone. An information-theoretic line of work provides the conceptual basis for our method. Predictive $\mathcal{V}$-usable information formalizes how much label-relevant information a model family can exploit under computational constraints, and its pointwise extension characterizes instance-level difficulty \citep{xu2020theoryusableinformationcomputational, pmlr-v162-ethayarajh22a}. Building on this perspective, \citet{kim-etal-2025-detecting} argue that hallucination is fundamentally a layer-wise information-deficiency phenomenon: usable information evolves non-monotonically across depth, so final-layer analysis can miss reliability-relevant gains and losses arising during intermediate computation. This line of work provides the closest conceptual basis for our method. \textbf{Our contribution} is to move from diagnosis to conformalization, using layer-wise internal information directly as the answer-level nonconformity score that drives prediction-set construction.

\section{Methodology}
\subsection{Preliminaries}
\label{sec:preliminaries}
We work in the standard split conformal prediction (SCP) setting for question answering. Let \(D_{\mathrm{cal}}=\{(x_i,y_i^*)\}_{i=1}^N
\) be a held-out calibration set, where \(x_i\in\mathcal X\) is the \(i\)-th question and \(y_i^*\in\mathcal Y\) its ground-truth answer. For each calibration question \(x_i\), we sample \(M\) responses from the deployed language model \(\mathcal M:\mathcal X\to\mathcal Y\), producing a candidate pool \(\{y_j^{(i)}\}_{j=1}^M\).

For multiple-choice QA, each sampled response is parsed into one answer option; for open-domain QA, sampled responses are grouped into semantic answer units following prior conformal QA protocols \citep{quach2024conformallanguagemodeling, su2024apienoughconformalprediction}. Let \(\mathcal A(x_i)\) denote the set of distinct candidate answer units induced by the sampled responses for \(x_i\), and define
\[
\mathcal A^*(x_i,y_i^*)
:=
\{a\in\mathcal A(x_i): a \text{ is admissible for } y_i^*\},
\]
the subset of admissible answer units in the sampled pool. Here admissibility is defined by exact match in MCQA and by semantic admissibility in open-domain QA. Let \(F(a;x_i)\) be any fixed answer-level reliability score, with larger values indicating more trustworthy answers. The corresponding calibration nonconformity score is
\begin{equation}
s_i(F)=
\begin{cases}
1-\max\limits_{a\in\mathcal A^*(x_i,y_i^*)}F(a;x_i),
& \text{if } \mathcal A^*(x_i,y_i^*)\neq\emptyset,\\[4pt]
\infty,
& \text{if } \mathcal A^*(x_i,y_i^*)=\emptyset.
\end{cases}
\label{eq:calibration-score}
\end{equation}
Given a target risk level \(\alpha\in(0,1)\), define the conformal threshold
\begin{equation}
\widehat q_\alpha(F)
:=
\operatorname{Quantile}\!\left(
1-\alpha;\,
\{s_i(F)\}_{i=1}^N \cup \{\infty\}
\right),
\label{eq:quantile}
\end{equation}
and the resulting prediction set for a new question \(x\) by
\begin{equation}
\widehat C_\alpha(x;F)
=
\left\{
a\in\mathcal A(x):
1-F(a;x)\le \widehat q_\alpha(F)
\right\}.
\label{eq:scp-set}
\end{equation}
Under exchangeability of the calibration and test examples, the resulting set predictor satisfies the standard marginal coverage guarantee
\begin{equation}
\mathbb P\!\left(
\widehat C_\alpha(X;F)\cap \mathcal A^*(X,Y^*)\neq\emptyset
\right)\ge 1-\alpha,
\label{eq:marginal-coverage}
\end{equation}
where \((X,Y^*)\) denotes a fresh test question-answer pair. Equivalently, with probability at least \(1-\alpha\), the conformal prediction set contains at least one answer unit admissible for the ground truth \citep{angelopoulos2022gentleintroductionconformalprediction}. Our contribution is thus a principled internal reliability score \(F\) that instantiates this otherwise standard conformal construction.

A practical complication in conformal QA is that candidate sets are formed from a finite number \(M\) of sampled responses. Consequently, some calibration questions may contain no admissible answer unit in the sampled pool. We retain such examples rather than filtering them out, since filtering changes the calibration distribution and can be especially harmful in cross-domain QA. This induces a minimum manageable risk level
\begin{equation}
\alpha_l
=
\frac{N}{N+1}\cdot
\frac{1}{N}
\sum_{i=1}^N
\mathbf 1\!\left\{
\mathcal A^*(x_i,y_i^*)=\emptyset
\right\},
\label{eq:minimum-risk}
\end{equation}
so practical guarantees based on sampled candidate pools are meaningful only when \(\alpha\ge \alpha_l\). Appendix~\ref{app:risk-floor} gives a short derivation and interpretation of this finite-sampling risk floor.

\subsection{Layer-wise Usable Information}
\label{sec:layerwise-ui}
We now define the internal reliability signal used by Layerwise CP. Our construction follows the layer-wise usable information view of \citet{kim-etal-2025-detecting}, adapted here from answerability analysis to conformal answer generation. For a question \(x\) and one sampled response \(y=(y_1,\ldots,y_T)\), let \(\mathcal L\) denote the set of transformer layers of the pretrained language model. For each layer \(\ell\in\mathcal L\), projecting the hidden state through the pretrained LM head induces next-token distributions both with question context, \(p_\ell(y_t\mid y_{<t},x)\), and with null context, \(p_\ell(y_t\mid y_{<t},\emptyset)\), where \(\emptyset\) denotes the absence of the question context.

We define the empirical predictive \(\ell\)-entropy of the realized response \(y\) by
\begin{equation}
H_\ell(y\mid x)
=
\frac{1}{T}\sum_{t=1}^T
-\log p_\ell(y_t\mid y_{<t},x),
\qquad
H_\ell(y\mid \emptyset)
=
\frac{1}{T}\sum_{t=1}^T
-\log p_\ell(y_t\mid y_{<t},\emptyset),
\label{eq:layer-entropy}
\end{equation}
and the corresponding per-layer usable information by
\begin{equation}
I_\ell(x\rightarrow y)
=
H_\ell(y\mid \emptyset)-H_\ell(y\mid x).
\label{eq:per-layer-information}
\end{equation}
Aggregating over a scored layer subset \(\mathcal L_s\subseteq \mathcal L\) gives the layer-wise usable information
\begin{equation}
LI(x\rightarrow y)
=
\sum_{\ell\in\mathcal L_s} I_\ell(x\rightarrow y),
\label{eq:total-li}
\end{equation}
with default choice \(\mathcal L_s=\mathcal L\). Positive \(I_\ell(x\rightarrow y)\) indicates that conditioning on the question reduces uncertainty for the realized response at layer \(\ell\), whereas negative values indicate the reverse. Because reliability-relevant evidence can vary non-monotonically across depth, Eq.~\eqref{eq:total-li} aggregates layer-wise gains and losses rather than relying on the final layer alone.

Because the absolute scale of \(LI(x\rightarrow y)\) can vary across questions and model backbones, we normalize it within each sampled candidate pool:
\begin{equation}
\widetilde{LI}(x_i\rightarrow y_j^{(i)})
=
\frac{
LI(x_i\rightarrow y_j^{(i)})
-
\min_{1\le m\le M} LI(x_i\rightarrow y_m^{(i)})
}{
\max_{1\le m\le M} LI(x_i\rightarrow y_m^{(i)})
-
\min_{1\le m\le M} LI(x_i\rightarrow y_m^{(i)})
+\varepsilon
},
\label{eq:normalized-li}
\end{equation}
where \(\varepsilon>0\) is a small constant. This preserves the within-question ranking of sampled responses while reducing scale variation across candidate pools. Appendix~\ref{app:usable-information} gives brief background on usable information and clarifies the interpretation of LI.

\subsection{Layer-wise Nonconformity Score}
\label{sec:layerwise-score}
The central step in Layerwise CP is to convert response-level layer-wise usable information into an answer-level reliability score. For each candidate answer unit \(a\in\mathcal A(x_i)\), let
\begin{equation}
\mathcal J_a^{(i)}
=
\{j\in\{1,\ldots,M\}: \mathrm{parse}(y_j^{(i)})=a\}
\label{eq:index-set}
\end{equation}
denote the set of sampled responses that induce answer \(a\). We first define the answer-frequency score
\begin{equation}
F_f(a;x_i)
=
\frac{|\mathcal J_a^{(i)}|}{M},
\label{eq:frequency-score}
\end{equation}
which measures cross-sample consensus and is widely used in self-consistency and conformal QA pipelines \citep{kuhn2023semanticuncertaintylinguisticinvariances, wang-etal-2025-sconu}. Since frequency alone remains an output-level statistic, we complement it with an internal support score obtained by averaging normalized LI over all sampled responses that map to the same answer:
\begin{equation}
F_{LI}(a;x_i)
=
\frac{1}{|\mathcal J_a^{(i)}|}
\sum_{j\in\mathcal J_a^{(i)}}
\widetilde{LI}(x_i\rightarrow y_j^{(i)}).
\label{eq:li-answer-score}
\end{equation}
This aggregation maps response-level internal evidence to the answer-unit level required by conformal QA and stabilizes the score when several generations parse to the same answer. This aggregation aligns with recent representation-level evidence aggregation, which shows that multiple generations supporting the same answer can provide stronger evidence than any single response alone \citep{jiang2025representationconsistencyaccuratecoherent}.

We then combine internal support and answer frequency through
\begin{equation}
F^{\mathrm{LW}}(a;x_i)
=
w_{LI}F_{LI}(a;x_i)+w_fF_f(a;x_i),
\qquad
w_{LI},w_f\ge 0,
\qquad
w_{LI}+w_f=1,
\label{eq:final-reliability}
\end{equation}
where \(F_{LI}(a;x_i)\) measures internal contextual support, while \(F_f(a;x_i)\) regularizes against response-level noise through cross-sample agreement. In our experiments, we set \(w_{LI}=w_f=0.5\), following \citet{jiang2025representationconsistencyaccuratecoherent}, where equal weighting provided a strong and stable aggregation rule. We emphasize, however, that Layerwise CP is defined for any fixed choice of \((w_{LI},w_f)\) specified before evaluation. Accordingly, our method modifies the reliability score within a fixed conformal construction, rather than the conformal wrapper itself.

Using \(F^{\mathrm{LW}}\), the Layerwise calibration nonconformity score becomes
\begin{equation}
s_i^{\mathrm{LW}}
=
s_i(F^{\mathrm{LW}})
=
\begin{cases}
1-\max\limits_{a\in\mathcal A^*(x_i,y_i^*)}F^{\mathrm{LW}}(a;x_i),
& \text{if } \mathcal A^*(x_i,y_i^*)\neq\emptyset,\\[4pt]
\infty,
& \text{if } \mathcal A^*(x_i,y_i^*)=\emptyset,
\end{cases}
\label{eq:lw-score}
\end{equation}
with corresponding conformal threshold
\begin{equation}
\widehat q_\alpha^{\mathrm{LW}}
=
\widehat q_\alpha(F^{\mathrm{LW}}),
\label{eq:lw-quantile}
\end{equation}
and test-time prediction set
\begin{equation}
\widehat C_\alpha^{\mathrm{LW}}(x)
=
\widehat C_\alpha(x;F^{\mathrm{LW}})
=
\left\{
a\in\mathcal A(x):
1-F^{\mathrm{LW}}(a;x)\le \widehat q_\alpha^{\mathrm{LW}}
\right\}.
\label{eq:lw-set}
\end{equation}
Since the conformal construction is unchanged, standard marginal validity is preserved; the difference lies in the ranking of candidate answers, and hence in the efficiency of the resulting prediction sets. Appendix~\ref{app:validity} gives the formal validity proof, and Appendix~\ref{app:algorithm} summarizes the implementation.

\subsection{Layer-wise Information under Domain Shift}
\label{sec:why-li-helps}
The relevant issue is therefore efficiency rather than validity. Within a fixed split conformal procedure, different nonconformity scores induce different orderings of candidate answers; at the same target risk, a score that better separates admissible from inadmissible answers should exclude spurious candidates earlier and thus return smaller, less diffuse prediction sets. This interpretation is also consistent with the information-theoretic view of conformal prediction, in which prediction-set size at controlled risk reflects residual uncertainty.

Layer-wise information is motivated by this ranking problem, especially under calibration--deployment mismatch. Standard uncertainty signals such as final-layer probabilities, predictive entropy, semantic entropy, or answer frequency are functions of the terminal output distribution alone. By contrast, \(LI(x\rightarrow y)\) aggregates how conditioning on the question changes predictive entropy throughout model depth. This distinction matters because usable information need not evolve monotonically across layers: intermediate computations may create, attenuate, or partially recover reliability-relevant evidence before the final output is formed. A score based only on the terminal distribution can therefore compress distinctions that remain visible in the internal trajectory, particularly when surface-level features become unreliable proxies under domain shift.

Our claim is consequently empirical and ranking-based. If the LI-based score in Eq.~\eqref{eq:final-reliability} preserves the admissible--inadmissible ordering more robustly than output-level baselines, then the resulting conformal sets should be tighter at comparable empirical validity. We therefore evaluate Layerwise CP through its EMR--APSS operating point, with lower APSS at similar EMR indicating improved efficiency. Appendix~\ref{app:itcp-layerwise} places this interpretation in an information-theoretic context, but we use that connection as an explanatory lens rather than as a proof that layer-wise scores must dominate final-output scores in all settings.

\section{Experiments}
\subsection{Experimental Settings}
Following \citet{wang-etal-2025-sconu}, we evaluate both in-domain and cross-domain question answering under a matched generation and calibration protocol. We consider two closed-ended benchmarks, \textbf{MMLU-Pro} \citep{NEURIPS2024_ad236edc} and \textbf{MedMCQA} \citep{pmlr-v174-pal22a}, and two open-domain benchmarks, \textbf{TriviaQA} \citep{joshi-etal-2017-triviaqa} and \textbf{CoQA} \citep{10.1162/tacl_a_00266}. For cross-domain evaluation, we adopt the subject-wise protocol on MMLU-Pro used in SConU, in which the calibration set is drawn from one discipline and the test set from another, thereby creating a controlled calibration--deployment mismatch. For open-domain QA, we follow the same validation-split sampling regime as SConU, using 4,000 randomly selected examples from TriviaQA and CoQA.

We use a five-model white-box LLM suite aligned with the model families studied in SConU and COIN: \textbf{Qwen-2.5-3B-Instruct}, \textbf{Qwen-2.5-7B-Instruct}, \textbf{LLaMA-3.1-8B-Instruct}, \textbf{Vicuna-13B-v1.5}, and \textbf{Qwen-2.5-14B-Instruct}. These models belong respectively to the Qwen2.5 \citep{yang2024qwen25mathtechnicalreportmathematical}, Llama 3.1 \citep{grattafiori2024llama3herdmodels}, and Vicuna \citep{vicuna2023} families. All models are used without fine-tuning. Because our method requires hidden states and intermediate logits, all experiments are conducted in a white-box inference setting.

Candidate generation follows the same sampling pipeline as SConU \citep{wang-etal-2025-sconu}. For four-option multiple-choice QA tasks, we sample \(M=20\) candidate responses per question; for MMLU-Pro, which expands most questions to ten answer choices, we set \(M=50\); and for TriviaQA and CoQA we sample \(M=10\) responses per prompt. We also follow the same prompting strategy: 3-shot prompts for MMLU-Pro and MedMCQA, and few-shot prompts for TriviaQA and CoQA. Decoding uses multinomial sampling with \texttt{do\_sample=True}, \texttt{num\_beams=1}, \texttt{top\_p=0.9}, and \texttt{temperature=1.0}; the maximum generation length is set to one token for multiple-choice tasks and 36 tokens for open-domain QA.

We compare our layer-wise nonconformity score against surface-level baselines under exactly the same candidate pools and conformal wrapper. For closed-ended QA, we include predictive-entropy and logit/frequency-based scores used in recent conformal QA pipelines \citep{kadavath2022languagemodelsmostlyknow, kumar2023conformalpredictionlargelanguage, su2024apienoughconformalprediction, wang-etal-2025-sconu}. For open-domain QA, we compare against semantic-entropy and self-consistency-style uncertainty scores, again keeping candidate generation and calibration fixed so that the only difference is the uncertainty signal being conformalized \citep{kuhn2023semanticuncertaintylinguisticinvariances, lin2024generatingconfidenceuncertaintyquantification, wang2024conuconformaluncertaintylarge}. This design isolates the contribution of internal layer-wise information from the effects of sampling or prompt design.

Unless otherwise stated, we use a calibration/test split ratio of \(0.5\) and repeat each experiment over 100 random trials. We report \textbf{empirical miscoverage rate} (EMR) for marginal validity, \textbf{average prediction set size} (APSS) for efficiency, and \textbf{size-stratified miscoverage} (SSM) as an approximate conditional-coverage diagnostic \citep{quach2024conformallanguagemodeling, wang-etal-2025-sconu}. For open-domain QA, correctness is evaluated using the same sentence-similarity-based protocol adopted in recent conformal QA work, with optional LLM-judgment checks in additional experiments \citep{su2024apienoughconformalprediction,wang-etal-2025-sconu,angelopoulos2026theoreticalfoundationsconformalprediction}. Since our information-theoretic interpretation is driven by conformal set size, APSS serves as the primary efficiency metric in the main text \citep{NEURIPS2024_b6fa3ed9}.

\subsection{Experimental Evaluation}
We evaluate Layerwise CP along three axes: calibration across target risk budgets, robustness under subject-wise cross-domain shift, and performance on single-domain and open-domain QA. Lower EMR, APSS, and SSM indicate better empirical operating points, with SSM used as a descriptive diagnostic rather than a formal conditional-coverage guarantee.

Table~\ref{Layerwise Performance} first shows that Layerwise CP behaves as expected across user-specified risk budgets. On both MMLU-Pro and MedMCQA, realized EMR increases monotonically with \(\beta \in \{0.1,0.2,0.3\}\) while remaining below the nominal level for all five backbones. Averaged across models, EMR on MMLU-Pro is \(0.087\), \(0.169\), and \(0.237\); on MedMCQA, it is \(0.089\), \(0.171\), and \(0.242\), at \(\beta=0.1,0.2,0.3\), respectively. This indicates that the layer-wise score remains well calibrated across operating points rather than only at a single risk level. We also observe a mild scaling trend: Qwen-2.5-14B-Instruct attains the lowest EMR at all three budgets on both datasets, consistent with stronger backbones inducing cleaner answer rankings under the same conformal pipeline.
\begin{table}[h]
\footnotesize
\caption{Performance across target risk budgets on MMLU-Pro and MedMCQA. Results report EMR at \(\beta \in \{0.1, 0.2, 0.3\}\) for five LLM backbones.}
\label{Layerwise Performance}
\begin{center}
\begin{tabular}{cccc}
\toprule
\multicolumn{1}{c}{\bf Dataset}  & \multicolumn{3}{c}{\bf MMLU-Pro (closed-ended)} \\
\midrule
\textbf{LLMs} / $\beta$ & \textbf{0.1} & \textbf{0.2} & \textbf{0.3} \\
Qwen-2.5-3B-Instruct  & $0.0901\pm0.0109$ & $0.1736\pm0.0117$ & $0.2455\pm0.0132$ \\
Qwen-2.5-7B-Instruct  & $0.0877\pm0.0086$ & $0.1698\pm0.0091$ & $0.2386\pm0.0106$ \\
Qwen-2.5-14B-Instruct & $0.0838\pm0.0074$ & $0.1662\pm0.0082$ & $0.2267\pm0.0095$ \\
LLaMA-3.1-8B-Instruct & $0.0860\pm0.0081$ & $0.1655\pm0.0089$ & $0.2324\pm0.0101$ \\
Vicuna-13B-v1.5       & $0.0885\pm0.0095$ & $0.1716\pm0.0102$ & $0.2410\pm0.0122$ \\
\midrule
\multicolumn{1}{c}{\bf Dataset}  & \multicolumn{3}{c}{\bf MedMCQA (closed-ended)} \\
\midrule
Qwen-2.5-3B-Instruct  & $0.0914\pm0.0082$ & $0.1762\pm0.0096$ & $0.2498\pm0.0111$ \\
Qwen-2.5-7B-Instruct  & $0.0889\pm0.0077$ & $0.1719\pm0.0084$ & $0.2422\pm0.0108$ \\
Qwen-2.5-14B-Instruct & $0.0849\pm0.0068$ & $0.1637\pm0.0073$ & $0.2341\pm0.0080$ \\
LLaMA-3.1-8B-Instruct & $0.0875\pm0.0073$ & $0.1680\pm0.0080$ & $0.2365\pm0.0096$ \\
Vicuna-13B-v1.5       & $0.0902\pm0.0081$ & $0.1747\pm0.0095$ & $0.2466\pm0.0110$ \\
\bottomrule
\end{tabular}
\end{center}
\end{table}

\begin{figure}[h]
\begin{center}
	\includegraphics[width=\textwidth]{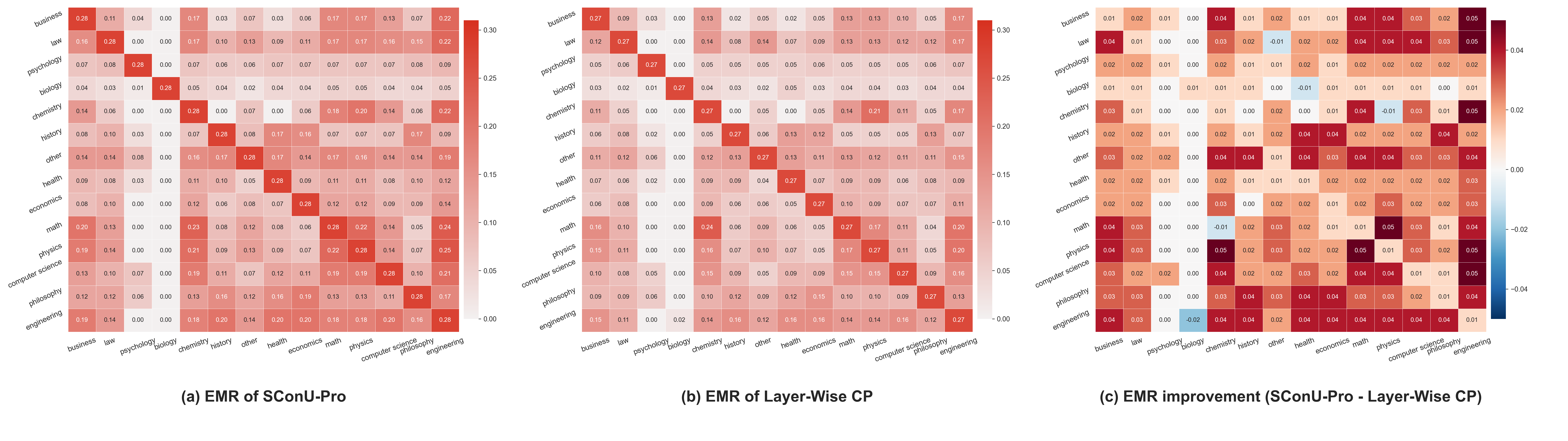}
\end{center}
\caption{Cross-domain EMR heatmaps on MMLU-Pro. Panels (a) and (b) show SConU-Pro and Layerwise CP, respectively; panel (c) reports the difference \(\mathrm{EMR}_{\text{SConU-Pro}}-\mathrm{EMR}_{\text{Layerwise}}\). Rows denote calibration domains and columns test domains. Lower EMR is better, so positive values in panel (c) indicate lower miscoverage for Layerwise CP, with broadly positive off-diagonal regions suggesting stronger cross-domain robustness.}
\label{EMRGraph}
\end{figure}

\begin{figure}[h]
\begin{center}
	\includegraphics[width=\textwidth]{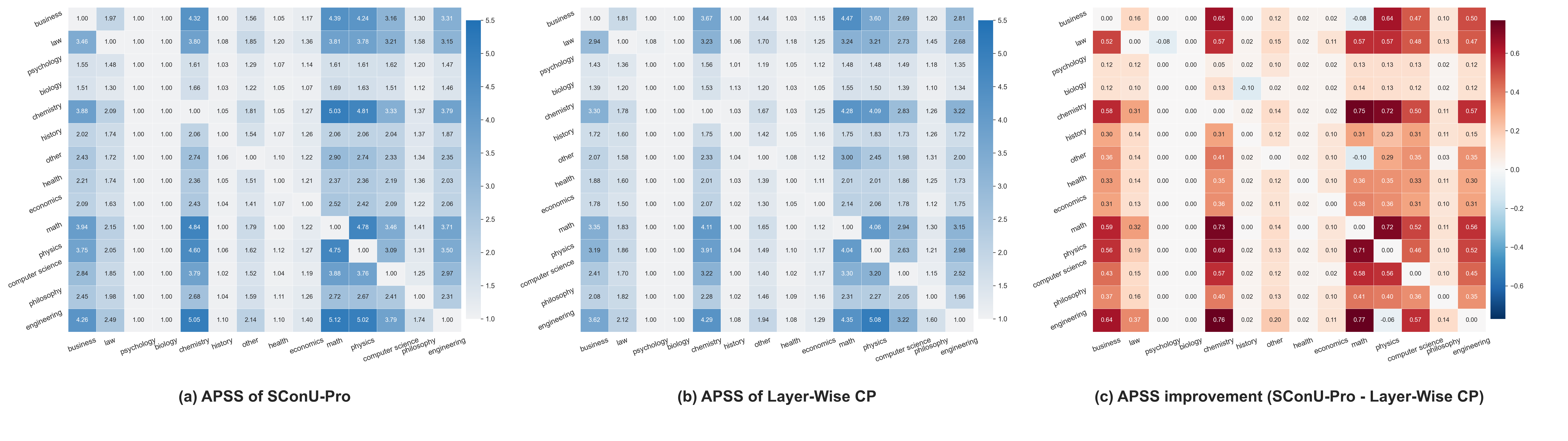}
\end{center}
\caption{Cross-domain APSS heatmaps on MMLU-Pro. Panels (a) and (b) show SConU-Pro and Layerwise CP, respectively; panel (c) reports the difference \(\mathrm{APSS}_{\text{SConU-Pro}}-\mathrm{APSS}_{\text{Layerwise}}\). Rows denote calibration domains and columns test domains. Lower APSS is better, so positive values in panel (c) indicate smaller and more efficient prediction sets for Layerwise CP, especially on harder cross-domain transfers.}
\label{APSSGraph}
\end{figure}

The main empirical gains appear in the subject-wise cross-domain setting. Figures~\ref{EMRGraph} and~\ref{APSSGraph} show that most off-diagonal transfers improve in EMR, while APSS gains are broad and especially pronounced on harder transfers involving math, physics, chemistry, and engineering. This pattern is consistent with the central hypothesis: layer-wise internal information becomes most useful when calibration and deployment domains are mismatched.

Table~\ref{Cross-domain Performance} summarizes this effect across five white-box LLMs. Relative to SConU-Pro, Layerwise CP improves both EMR and APSS for all five models. Averaged over models, EMR decreases from \(0.144\) to \(0.119\), an absolute reduction of \(0.025\) (\(17.6\%\) relative), while APSS decreases from \(2.172\) to \(1.804\), an absolute reduction of \(0.368\) (\(16.9\%\) relative). SSM also improves on average, from \(0.213\) to \(0.194\), although this improvement is not uniform: four of the five models improve, while Vicuna-13B shows a slight regression from \(0.218\) to \(0.221\). The strongest overall backbone is Qwen-2.5-14B-Instruct, which achieves \(0.110\) EMR, \(1.68\) APSS, and \(0.178\) SSM under Layerwise CP. Just as importantly, the standard deviations of EMR and APSS are consistently smaller than those of SConU-Pro, indicating that the improvements are stable across random trials rather than driven by a few favorable splits.

\begin{table}[h]
\footnotesize
\caption{Cross-domain MMLU-Pro summary comparing SConU-Pro and Layerwise CP. We report EMR mean/std, APSS mean/std, and size-stratified miscoverage (SSM). Lower EMR, APSS, and SSM indicate better empirical operating points.}
\label{Cross-domain Performance}
\begin{center}
\begin{tabular}{ccccc}
\toprule
\multicolumn{1}{c}{\bf LLMs} & \multicolumn{1}{c}{\bf Method} & \multicolumn{1}{c}{\bf EMR Mean/Std} & \multicolumn{1}{c}{\bf APSS Mean/Std} & \multicolumn{1}{c}{\bf SSM} \\
\midrule
\multirow{2}{*}{Qwen-2.5-3B} & SConU-Pro     & $0.153/0.011$ & $2.31/0.14$ & $0.221$ \\
& Layerwise CP & $0.128/0.009$ & $1.91/0.12$ & $0.197$ \\
\midrule
\multirow{2}{*}{Qwen-2.5-7B} & SConU-Pro     & $0.145/0.010$ & $2.18/0.13$ & $0.214$ \\
& Layerwise CP & $0.120/0.008$ & $1.82/0.11$ & $0.189$ \\
\midrule
\multirow{2}{*}{Qwen-2.5-14B} & SConU-Pro    & $0.136/0.009$ & $2.03/0.11$ & $0.203$ \\
& Layerwise CP & $0.110/0.007$ & $1.68/0.09$ & $0.178$ \\
\midrule
\multirow{2}{*}{LLaMA-3.1-8B} & SConU-Pro    & $0.140/0.009$ & $2.10/0.12$ & $0.208$ \\
& Layerwise CP & $0.114/0.007$ & $1.74/0.10$ & $0.183$ \\
\midrule
\multirow{2}{*}{Vicuna-13B} & SConU-Pro    & $0.148/0.010$ & $2.24/0.13$ & $0.218$ \\
& Layerwise CP & $0.123/0.008$ & $1.87/0.11$ & $0.221$ \\
\bottomrule
\end{tabular}
\end{center}
\end{table}

\begin{table}[h]
\footnotesize
\caption{Open-domain QA evaluation on TriviaQA and CoQA, comparing Semantic-Entropy, SConU-Pro, and Layerwise CP. Entries are reported as TriviaQA / CoQA for each metric.}
\label{Open-domain Performance}
\begin{center}
\begin{tabular}{ccccc}
\toprule
\multicolumn{2}{c}{\bf Dataset} & \multicolumn{3}{c}{\bf TriviaQA / CoQA} \\
\midrule
\multicolumn{1}{c}{\bf LLMs} & \multicolumn{1}{c}{\bf Method} & \multicolumn{1}{c}{\bf EMR@0.3} & \multicolumn{1}{c}{\bf APSS} & \multicolumn{1}{c}{\bf SSM} \\
\midrule
\multirow{3}{*}{Qwen-2.5-7B} & Semantic-Entropy & $0.229/0.238$ & $2.46/2.58$ & $0.341/0.352$ \\
& SConU-Pro    & $0.215/0.223$ & $2.30/2.42$ & $0.329/0.338$ \\
& Layerwise CP & $0.192/0.226$ & $1.98/2.09$ & $0.301/0.340$ \\
\midrule
\multirow{3}{*}{LLaMA-3.1-8B} & Semantic-Entropy & $0.224/0.234$ & $2.39/2.50$ & $0.335/0.346$ \\
& SConU-Pro    & $0.209/0.220$ & $2.24/2.36$ & $0.321/0.333$ \\
& Layerwise CP & $0.187/0.221$ & $1.92/2.03$ & $0.295/0.336$ \\
\bottomrule
\end{tabular}
\end{center}
\end{table}

Table~\ref{Open-domain Performance} shows a consistent but dataset-dependent benefit of layer-wise scoring in open-domain QA. On TriviaQA, Layerwise CP improves all three metrics over both Semantic-Entropy and SConU-Pro for both backbones, reducing EMR from \(0.215\) to \(0.192\) and APSS from \(2.30\) to \(1.98\) on Qwen-2.5-7B, and from \(0.209\) to \(0.187\) and \(2.24\) to \(1.92\), respectively, on LLaMA-3.1-8B. On CoQA, the gains are more concentrated in efficiency: relative to SConU-Pro, APSS drops from \(2.42\) to \(2.09\) on Qwen-2.5-7B and from \(2.36\) to \(2.03\) on LLaMA-3.1-8B, while EMR and SSM remain broadly comparable. Taken together, these results indicate that layer-wise scores reliably tighten conformal sets in open-domain QA, with the clearest joint gains in both validity and efficiency appearing on TriviaQA.

\begin{wraptable}{r}{0.65\textwidth}
\centering
\footnotesize
\caption{Single-domain MedMCQA results at calibration ratio \(0.35\), comparing Entropy, SConU-Pro, and Layerwise CP across two models via EMR@0.3, APSS, and SSM.}
\label{Single-domain Performance}
\begin{tabular}{ccccc}
\toprule
\multicolumn{1}{c}{\bf LLMs}  & \multicolumn{1}{c}{\bf Method} & \multicolumn{1}{c}{\bf EMR@0.3} & \multicolumn{1}{c}{\bf APSS} & \multicolumn{1}{c}{\bf SSM} \\
\midrule
\multirow{3}{*}{Qwen-2.5-7B} & Entropy      & $0.246$ & $1.84$ & $0.328$ \\
& SConU-Pro    & $0.232$ & $1.70$ & $0.313$ \\
& Layerwise CP & $0.226$ & $1.63$ & $0.308$ \\
\midrule
\multirow{3}{*}{LLaMA-3.1-8B} & Entropy      & $0.240$ & $1.76$ & $0.322$ \\
& SConU-Pro    & $0.227$ & $1.64$ & $0.309$ \\
& Layerwise CP & $0.221$ & $1.58$ & $0.311$ \\
\bottomrule
\end{tabular}
\end{wraptable}

Table~\ref{Single-domain Performance} shows a milder but still meaningful advantage in the single-domain MedMCQA setting. Compared with the entropy baseline, Layerwise CP improves all three metrics on both models. Relative to SConU-Pro, it further reduces EMR and APSS on both backbones, while SSM remains broadly comparable, with a slight gain on Qwen-2.5-7B and near parity on LLaMA-3.1-8B. This suggests that layer-wise scoring remains useful in-domain, although the margin is naturally smaller than under cross-domain shift.

Overall, the experiments show that Layerwise CP remains well behaved across target risk budgets, delivers its largest gains under subject-wise cross-domain shift, and consistently reduces APSS. The strongest empirical result is on cross-domain MMLU-Pro, where Layerwise CP improves both EMR and APSS relative to SConU-Pro across all five backbones. In single-domain and open-domain QA, the method remains competitive and often improves efficiency, with the clearest additional gains on TriviaQA. Taken together, these results support layer-wise internal trajectories as a more informative conformal scoring signal than surface-based uncertainty, especially when calibration and deployment domains differ.

\section{Conclusion}
We presented \textit{Layerwise CP}, a conformal prediction framework for LLM question answering that uses layer-wise usable information as the nonconformity score within an otherwise standard split conformal pipeline. Across closed-ended and open-domain QA benchmarks, the method remains well behaved across risk budgets and achieves its clearest improvements under cross-domain shift, where it consistently improves empirical EMR--APSS operating points relative to strong surface-based baselines. Overall, the benefit of layer-wise scoring is that it provides a stronger answer-ranking signal within the same procedure.

More broadly, the results suggest that internal representations can serve as a useful basis for conformal scoring in LLMs, particularly when output-level uncertainty becomes unstable under calibration--deployment mismatch. At the same time, our contribution is empirical and score-level rather than a new validity theorem under shift. An important direction for future work is therefore to make this connection more precise theoretically, especially by relating layer-wise information to ranking quality, conditional entropy, and the efficiency of conformal prediction sets. Clarifying these links would help explain more formally when and why internal scores yield tighter conformal sets, and would further connect mechanistic analysis of LLM computation with uncertainty quantification under distribution shift.

\bibliography{colm2026_conference}
\bibliographystyle{colm2026_conference}

\newpage
\appendix

\section{Finite-Sampling Risk Floor for Split Conformal QA}
\label{app:risk-floor}
This appendix justifies the minimum manageable risk level in Eq.~\eqref{eq:minimum-risk}. Since the standard split conformal construction is already given in Sec.~\ref{sec:preliminaries}, we focus only on the additional issue created by finite candidate pools. For any confidence threshold \(\lambda\in[0,1]\), define the reliable response set
\begin{equation}
C_{\lambda}(x_i)=\{a\in\mathcal A(x_i): F(a;x_i)\ge 1-\lambda\},
\label{eq:app-reliable-set}
\end{equation}
and the corresponding miscoverage loss
\begin{equation}
l\!\left(C_{\lambda}(x_i),y_i^*\right)=\mathbf 1\{y_i^*\notin C_{\lambda}(x_i)\}.
\label{eq:app-miscoverage-loss}
\end{equation}
The empirical miscoverage rate on the calibration set is then
\begin{equation}
L_N(\lambda)=\frac{1}{N}\sum_{i=1}^N l\!\left(C_{\lambda}(x_i),y_i^*\right).
\label{eq:app-empirical-miscoverage}
\end{equation}

\begin{proposition}[Finite-sampling risk floor]
\label{prop:risk-floor}
For any \(\lambda\in[0,1]\),
\begin{equation}
l\!\left(C_{\lambda}(x_i),y_i^*\right)\ge \mathbf 1\{y_i^*\notin \mathcal A(x_i)\},
\label{eq:app-loss-lower-bound}
\end{equation}
and therefore
\begin{equation}
L_N(\lambda)\ge \frac{1}{N}\sum_{i=1}^N \mathbf 1\{y_i^*\notin \mathcal A(x_i)\}.
\label{eq:app-miscoverage-lower-bound}
\end{equation}
Equivalently, the sampled split-conformal procedure has minimum manageable risk level
\begin{equation}
\alpha_l=\frac{N}{N+1}\cdot \frac{1}{N}\sum_{i=1}^N \mathbf 1\{y_i^*\notin \mathcal A(x_i)\}.
\label{eq:app-minimum-risk}
\end{equation}
\end{proposition}

\begin{proof}
By construction, \(C_{\lambda}(x_i)\subseteq \mathcal A(x_i)\) for every \(\lambda\in[0,1]\). Hence, if no admissible answer appears in the sampled candidate pool, i.e.,
\[
y_i^*\notin \mathcal A(x_i),
\]
then \(y_i^*\notin C_{\lambda}(x_i)\) for every \(\lambda\). This proves Eq.~\eqref{eq:app-loss-lower-bound}. Averaging over \(i=1,\dots,N\) gives Eq.~\eqref{eq:app-miscoverage-lower-bound}. The expression in Eq.~\eqref{eq:app-minimum-risk} is the same lower bound written in the finite-sample normalization induced by the split-conformal quantile rule.
\end{proof}

Proposition~\ref{prop:risk-floor} formalizes a simple but important limitation: if the sampled candidate pool misses an admissible answer, then no conformal threshold can recover coverage for that example. Thus, risk levels below \(\alpha_l\) are unattainable unless one changes the sampling budget \(M\), the decoding scheme, or the answer aggregation rule. This is why we keep all calibration examples, including those whose sampled candidate pools contain no admissible answer, rather than removing them.

\section{Usable Information}
\label{app:usable-information}
This appendix gives the background for the internal score used in Layerwise CP. Let \(\mathcal V\) denote a constrained predictive model family. Predictive \(\mathcal V\)-usable information measures how much access to the input \(X\) reduces the best achievable predictive entropy of \(Y\) relative to a null input:
\begin{equation}
I_{\mathcal V}(X \rightarrow Y)
=
H_{\mathcal V}(Y \mid \emptyset)-H_{\mathcal V}(Y \mid X),
\label{eq:app-v-info}
\end{equation}
where \(H_{\mathcal V}(Y\mid X)\) is the minimum expected log-loss attainable by models in \(\mathcal V\) when predicting \(Y\) from \(X\), and \(H_{\mathcal V}(Y\mid \emptyset)\) is the analogous quantity without access to the input \citep{xu2020theoryusableinformationcomputational}. Its pointwise extension refines this idea to individual examples, quantifying how much a specific input-output pair benefits from access to the input under the same computational constraints \citep{pmlr-v162-ethayarajh22a}.

Our LI score instantiates this perspective inside a pretrained autoregressive language model. Rather than fitting an auxiliary predictor family, we use the pretrained LM head to probe the token distribution induced by each internal layer. For a question \(x\) and sampled response \(y\), Sec.~\ref{sec:layerwise-ui} defines a per-layer information term \(I_{\ell}(x\rightarrow y)\) by comparing the token-level predictive entropy with and without the question context, and then sums these contributions across a scored layer set \(L_s\).

This layer-wise aggregation is important because usable information in LLMs is generally non-monotonic across depth: reliability-relevant evidence can be gained, attenuated, or partially recovered before the final output is formed \citep{kim-etal-2025-detecting}. A final-layer-only statistic therefore need not capture the full internal trajectory associated with a response. In our setting, LI is not used as a post hoc probing quantity, but as the internal reliability signal that is later conformalized.

Finally, the normalization in Eq.~\eqref{eq:normalized-li} is purely a within-pool rescaling step. It preserves the ranking of sampled responses for the same question while reducing scale variation across questions and backbones, making the downstream answer-level aggregation more stable.

\section{Marginal Validity of Layerwise CP}
\label{app:validity}
\begin{proposition}[Marginal validity of Layerwise CP]
\label{prop:app-validity}
Let \(D_{\mathrm{cal}}=\{(x_i,y_i^*)\}_{i=1}^{N}\) be a calibration set and let \((x_{N+1},y_{N+1}^*)\) be a fresh test sample. Assume that
\[
(x_1,y_1^*),\ldots,(x_N,y_N^*),(x_{N+1},y_{N+1}^*)
\]
are exchangeable. For each example, candidate generation, parsing, LI computation, and score construction are performed by the same randomized procedure, independently across examples. Define
\begin{equation}
s_i=
\begin{cases}
1-\max\limits_{a\in\mathcal A(x_i):\,a\asymp y_i^*}F(a;x_i), & \text{if }\mathcal A^*(x_i,y_i^*)\neq\emptyset,\\[4pt]
\infty, & \text{otherwise,}
\end{cases}
\label{eq:app-score}
\end{equation}
let
\begin{equation}
\widehat q_{\alpha}=\mathrm{Quantile}\!\left(1-\alpha;\{s_i\}_{i=1}^{N}\cup\{\infty\}\right),
\label{eq:app-quantile}
\end{equation}
and define
\begin{equation}
\widehat C_{\alpha}^{\mathrm{LW}}(x)=\{a\in\mathcal A(x):1-F(a;x)\le \widehat q_{\alpha}\}.
\label{eq:app-set}
\end{equation}
Then
\begin{equation}
\mathbb P\!\left(\widehat C_{\alpha}^{\mathrm{LW}}(x_{N+1})\cap \mathcal A^*(x_{N+1},y_{N+1}^*)\neq\emptyset\right)\ge 1-\alpha.
\label{eq:app-validity-result}
\end{equation}
\end{proposition}

\begin{proof}
Let
\[
Z_i=(x_i,y_i^*,U_i),
\]
where \(U_i\) collects all auxiliary randomness used for candidate generation, decoding, parsing, and score computation on the \(i\)-th example. By assumption, the tuples \(Z_1,\ldots,Z_N,Z_{N+1}\) are exchangeable. Since the mapping \(Z_i\mapsto s_i\) is the same for all \(i\), the induced scores \(s_1,\ldots,s_N,s_{N+1}\) are also exchangeable.

Let
\[
k=\left\lceil (N+1)(1-\alpha)\right\rceil.
\]
By construction, \(\widehat q_{\alpha}\) is the \(k\)-th order statistic of \(s_1,\ldots,s_N,\infty\), which is the standard split-conformal threshold. Hence
\begin{equation}
\mathbb P(s_{N+1}\le \widehat q_{\alpha})\ge \frac{k}{N+1}\ge 1-\alpha.
\label{eq:app-rank}
\end{equation}

It remains to identify the event \(s_{N+1}\le \widehat q_{\alpha}\). By Eq.~\eqref{eq:app-score},
\[
s_{N+1}\le \widehat q_{\alpha}
\quad\Longleftrightarrow\quad
\exists\, a\in\mathcal A(x_{N+1}) \text{ such that } a\asymp y_{N+1}^*
\text{ and } 1-F(a;x_{N+1})\le \widehat q_{\alpha}.
\]
By Eq.~\eqref{eq:app-set}, this is exactly the event
\[
\widehat C_{\alpha}^{\mathrm{LW}}(x_{N+1})\cap \mathcal A^*(x_{N+1},y_{N+1}^*)\neq\emptyset.
\]
Combining this equivalence with Eq.~\eqref{eq:app-rank} proves Eq.~\eqref{eq:app-validity-result}.
\end{proof}

\section{Algorithmic Summary of Layerwise CP}
\label{app:algorithm}
Algorithm~\ref{alg:layerwise-cp} summarizes the full procedure used in our experiments.

\begin{algorithm}[h]
\caption{Layerwise CP}
\label{alg:layerwise-cp}
\begin{algorithmic}[1]
\Require Calibration set \(D_{\mathrm{cal}}=\{(x_i,y_i^*)\}_{i=1}^N\), sample size \(M\), scored layer set \(L_s\), normalization constant \(\varepsilon\), target risk level \(\alpha\)
\Ensure Split conformal threshold \(\widehat q_\alpha\) and prediction rule \(x\mapsto \widehat C_\alpha^{\mathrm{LW}}(x)\)

\Function{AnswerScore}{$x$}
    \State Sample \(M\) responses \(\{y_j\}_{j=1}^M\) from the deployed LLM
    \For{$j=1,\dots,M$}
        \State Compute \(LI(x\rightarrow y_j)\) using Eqs.~\eqref{eq:layer-entropy}--\eqref{eq:total-li}
    \EndFor
    \State Normalize \(\{LI(x\rightarrow y_j)\}_{j=1}^M\) within the sampled pool using Eq.~\eqref{eq:normalized-li}
    \State Parse \(\{y_j\}_{j=1}^M\) into distinct answer units \(\mathcal A(x)\)
    \For{each \(a\in\mathcal A(x)\)}
        \State Form \(\mathcal J_a=\{j:\mathrm{parse}(y_j)=a\}\)
        \State Compute \(F_f(a;x)\) by Eq.~\eqref{eq:frequency-score}
        \State Compute \(F_{LI}(a;x)\) by Eq.~\eqref{eq:li-answer-score}
        \State Set \(F(a;x)=\frac{1}{2}F_{LI}(a;x)+\frac{1}{2}F_f(a;x)\)
    \EndFor
    \State \Return \(\mathcal A(x)\) and \(\{F(a;x):a\in\mathcal A(x)\}\)
\EndFunction

\For{$i=1,\dots,N$}
    \State \((\mathcal A(x_i),F(\cdot;x_i))\gets\Call{AnswerScore}{x_i}\)
    \If{\(\mathcal A^*(x_i,y_i^*)\neq\emptyset\)}
        \State \(s_i\gets 1-\max_{a\in\mathcal A(x_i):\,a\asymp y_i^*}F(a;x_i)\)
    \Else
        \State \(s_i\gets \infty\)
    \EndIf
\EndFor

\State \(\widehat q_\alpha\gets \mathrm{Quantile}\!\left(1-\alpha;\{s_i\}_{i=1}^N\cup\{\infty\}\right)\)

\State For a new question \(x\), compute \((\mathcal A(x),F(\cdot;x))\gets\Call{AnswerScore}{x}\) and return
\[
\widehat C_\alpha^{\mathrm{LW}}(x)=\{a\in\mathcal A(x):1-F(a;x)\le \widehat q_\alpha\}.
\]
\end{algorithmic}
\end{algorithm}

\section{Information-Theoretic Interpretation of Layerwise CP}
\label{app:itcp-layerwise}

This appendix clarifies how the information-theoretic view of conformal prediction relates to Layerwise CP. The purpose is interpretive rather than theorem-proving: we use existing results on conformal prediction and conditional entropy to explain why improved EMR--APSS operating points, especially lower APSS at comparable EMR, are a meaningful empirical signature of a better nonconformity score.

\subsection{Conformal set size as an efficiency notion}

\citet{NEURIPS2024_b6fa3ed9} show that conformal prediction can be viewed as a form of variable-size list decoding and use this connection to upper bound the intrinsic uncertainty \(H(Y \mid X)\). In particular, for a conformal predictor \(C(X)\) with target risk \(\alpha \in (0,0.5)\), their simple Fano bound implies
\begin{equation}
H(Y \mid X)
\le
h_b(\alpha)
+
\alpha\,\mathbb E\!\left[
\log\!\bigl(|\mathcal Y|-|C(X)|\bigr)
\,\middle|\,
Y \notin C(X)
\right]
+
(1-\alpha_N)\,\mathbb E\!\left[
\log |C(X)|
\,\middle|\,
Y \in C(X)
\right],
\label{eq:app-simple-fano}
\end{equation}
where \(\alpha_N = \alpha - \frac{1}{N+1}\) and \(h_b(\cdot)\) is the binary entropy function. The key point for our purposes is that prediction-set size enters the bound directly. Thus, when realized miscoverage is controlled or comparable, smaller returned sets correspond to a tighter upper bound on residual uncertainty.

This is why set size is not merely a cosmetic quantity in conformal prediction. It is a formal notion of inefficiency, and it provides a principled reason to evaluate conformal predictors not only by coverage or miscoverage, but also by how concentrated their returned sets are. In our paper, APSS serves exactly this role.

\subsection{Connection to Layerwise CP}

Layerwise CP does not modify the split conformal wrapper and therefore does not change the underlying coverage theorem. Its only intervention is to replace the answer-ranking signal inside the wrapper. The relevant question is therefore not whether validity changes, but whether the new score yields better empirical operating points: in particular, whether it produces smaller prediction sets at comparable or lower realized miscoverage.

Let \(C^{\mathrm{base}}(X)\) and \(C^{\mathrm{LW}}(X)\) denote the prediction sets produced by a baseline score and by Layerwise CP, respectively, under the same nominal risk level. If Layerwise CP yields lower APSS while maintaining similar or lower EMR, then Eq.~\eqref{eq:app-simple-fano} gives a principled interpretation of that gain: the conformal predictor leaves less residual ambiguity over candidate answers while preserving the relevant notion of reliability.

In this sense, lower APSS is not just smaller output. It indicates that, after conditioning on the question, the conformal predictor needs to retain fewer candidate answer units to preserve correctness coverage. The returned semantic decision is therefore less diffuse. This is the information-theoretic sense in which improved answer ranking can produce more concentrated answer generation.

\subsection{Why this matters especially under domain shift}

Our central hypothesis is that layer-wise information improves ranking quality under calibration--deployment mismatch. Output-level uncertainty signals depend only on the final output distribution, whereas LI aggregates how the question context changes predictive uncertainty throughout the full computation. If admissible and inadmissible answers become harder to separate from surface statistics alone under domain shift, then a score built from internal trajectories can remain more stable. When this happens, the conformal predictor needs to include fewer spurious candidates to retain admissible ones, which appears empirically as lower APSS at comparable EMR, and in the strongest case as joint improvement in both metrics.

This is the main role of the information-theoretic perspective in our paper. We do \emph{not} claim a new theorem showing that LI universally dominates output-level scores, nor do we claim that Eq.~\eqref{eq:app-simple-fano} itself proves superiority under shift. Rather, we use it as a principled efficiency lens: once smaller conformal sets are observed at controlled or comparable realized risk, the corresponding upper bound on residual candidate-space uncertainty becomes tighter.

\subsection{Scope of the interpretation}

The simple Fano bound is stated for a conformal predictor over a fixed target space \(\mathcal Y\). Our QA setting instead works with per-question candidate answer units \(\mathcal A(x)\), induced by finite sampled response pools and answer aggregation. We therefore use the result as an interpretation of the \emph{effective} candidate space on which conformal prediction operates in practice, rather than as a new theorem specialized to our answer-unit construction.

This distinction matters for two reasons. First, the formal guarantee in our paper still comes from standard split conformal prediction. Second, our experiments do not directly estimate \(H(Y \mid X)\); rather, they test whether a new nonconformity score improves empirical EMR--APSS operating points. The information-theoretic result explains why such improvements are meaningful, but it is not itself the source of the coverage guarantee.

\subsection{Empirical operating-point view}

Figure~\ref{APSSvsEMR} provides a compact summary of this interpretation on cross-domain MMLU-Pro. For each backbone and each risk budget \(\beta \in \{0.1, 0.2, 0.3\}\), we plot one SConU-Pro operating point and one Layerwise CP operating point in the EMR--APSS plane and connect the pair by a line segment. Colors identify backbones, marker shapes identify risk budgets, and marker fill distinguishes the method.

Points closer to the lower-left correspond to better empirical operating points: lower realized miscoverage together with smaller prediction sets. The dominant visual pattern in Fig.~\ref{APSSvsEMR} is a shift from the SConU-Pro point toward the Layerwise CP point in that direction, indicating that the proposed layer-wise score typically returns tighter sets while also improving, or at least preserving, reliability under cross-domain mismatch. This figure is not intended as a proof of a new theorem. Its role is to make the ranking-based interpretation of Sec.~\ref{sec:why-li-helps} visually explicit: when LI improves answer ordering under shift, the conformal predictor moves toward smaller and more reliable returned sets.

\begin{figure}[h]
\begin{center}
    \includegraphics[width=\textwidth]{Graphs/APSSvsEMR.png}
\end{center}
\caption{EMR--APSS operating points on cross-domain MMLU-Pro across \(\beta \in \{0.1,0.2,0.3\}\). For each model and budget, a line segment connects the SConU-Pro point to the corresponding Layerwise CP point. Colors denote backbones, marker shapes denote risk budgets, and filled versus hollow markers denote SConU-Pro versus Layerwise CP. Points closer to the lower-left indicate smaller prediction sets at lower realized miscoverage. The prevailing down-left shift summarizes the main empirical pattern of Layerwise CP under domain shift.}
\label{APSSvsEMR}
\end{figure}

\section{Additional Cross-Domain Heatmaps}
\label{app:additional-heatmaps}
\begin{sidewaysfigure}
\centering
\begin{subfigure}{0.85\textheight}
    \centering
    \includegraphics[width=\linewidth]{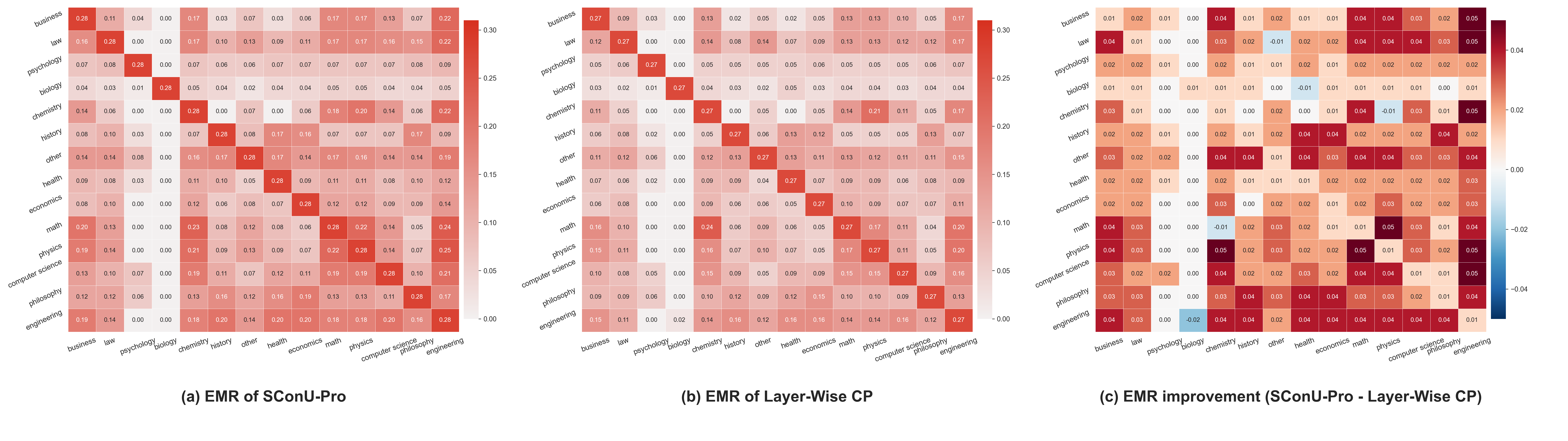}
    \caption{Cross-domain EMR on MMLU-Pro.}
    \label{app:EMRGraphLarge}
\end{subfigure}
\begin{subfigure}{0.85\textheight}
    \centering
    \includegraphics[width=\linewidth]{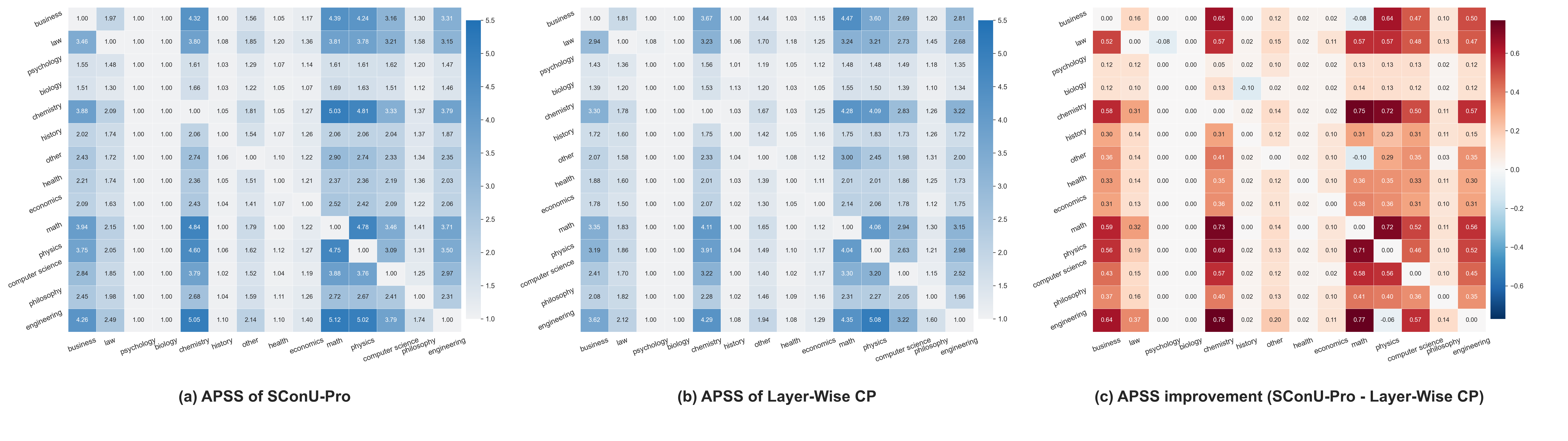}
    \caption{Cross-domain APSS on MMLU-Pro.}
    \label{app:APSSGraphLarge}
\end{subfigure}
\caption{Enlarged cross-domain heatmaps for readability. Left: EMR comparison between SConU-Pro and Layerwise CP, with panel (c) showing \(\mathrm{EMR}_{\text{SConU-Pro}}-\mathrm{EMR}_{\text{Layerwise}}\). Right: APSS comparison, with panel (c) showing \(\mathrm{APSS}_{\text{SConU-Pro}}-\mathrm{APSS}_{\text{Layerwise}}\). Rows denote calibration domains and columns denote test domains.}
\label{app:crossdomain-heatmaps}
\end{sidewaysfigure}

\end{document}